\documentclass{article} 
\usepackage{nips15submit_e,times}
\usepackage{hyperref}
\usepackage{url}
\usepackage{amssymb,amsfonts,amsmath,amsthm}
\usepackage[pdftex]{graphicx}
\usepackage{fancyhdr}

\newcommand{\gradtheta}{\nabla_{\theta}}

\newcommand{\argmax}{\operatornamewithlimits{argmax}}
\newcommand{\dkl}[2]{D_{\text{KL}}\left(#1||#2 \right)}

\newcommand{\feobj}{\mathcal F}
\newcommand{\rdisobj}{\mathcal J}

\newcommand{\setX}{\mathcal{X}}
\newcommand{\setY}{\mathcal{Y}}
\newcommand{\pbf}{\mathbf{p}}
\newcommand{\xbf}{\mathbf{x}}
\newcommand{\thetabf}{\boldsymbol{\theta}}
\newcommand{\reals}{\mathbb{R}}

\title{Adaptive information-theoretic bounded rational decision-making with parametric priors }

\author{
Jordi Grau-Moya  \\
\small Max Planck Institute for Intelligent Systems\\
\small Max Planck Institute for Biological Cybernetics\\
\small Spemannstra\ss e 38, T\"ubinben 72076, Germany \\
\small \texttt{jordi.grau@tuebingen.mpg.de} 
\And
Daniel A. Braun\\
\small Max Planck Institute for Intelligent Systems\\
\small Max Planck Institute for Biological Cybernetics\\
\small Spemannstra\ss e 38, T\"ubinben 72076, Germany \\
\small \texttt{daniel.braun@tuebingen.mpg.de} \\
}

\nipsfinalcopy 

\begin{document}

\maketitle

\section{Abstract}
\let\thefootnote\relax\footnote{Workshop on Bounded Optimality and Rational Metareasoning at Neural Information Processing Systems conference, Montreal, Canada,  2015}

Deviations from rational decision-making due to limited computational resources have been studied in the field of bounded rationality, originally proposed by Herbert Simon~\cite{simon1955behavioral}. There have been a number of different approaches to model bounded rationality~\cite{gershman2015computational} ranging from optimality principles \cite{Russell1995} to heuristics~\cite{gigerenzer1996reasoning}. Here we take an information-theoretic approach to bounded rationality~\cite{Ortega2013}, where information-processing costs are measured by the relative entropy between a posterior decision strategy and a given fixed prior strategy. In the case of multiple environments, it can be shown that there is an optimal prior rendering the bounded rationality problem equivalent to the rate distortion problem for lossy compression in information theory \cite{genewein2015}. Accordingly, the optimal prior and posterior strategies can be computed by the well-known Blahut-Arimoto algorithm which requires the computation of partition sums over all possible outcomes and cannot be applied straightforwardly to continuous problems.
Here we derive a sampling-based alternative update rule for the adaptation of prior behaviors of decision-makers and  we show convergence to the optimal prior predicted by rate distortion theory. Importantly, the update rule avoids typical infeasible operations such as the computation of partition sums. We show in simulations a proof of concept for discrete action and environment domains. This approach is not only interesting as a generic computational method, but might also provide a more realistic model of human decision-making processes occurring on a fast and a slow time scale.

\section{Information-theoretic bounded rationality}
In decision theory an agent faces an environment $y$ from the environment set $\mathcal Y$ and has to decide which action $x$ to take from the action set $\mathcal X$. The desirability of the action in that particular environment is quantified by the utility function $U : \setX \times \setY \mapsto \reals $. The objective of the decision-maker is to maximize the utility depending on the environment $y$, that is  
\begin{equation}
x^*(y) = \argmax_x U(x,y). \nonumber
\end{equation}
When the number of possible actions is vast, the maximization problem for any environment becomes computationally infeasible because the decision-maker must evaluate the utility of every possible action and choose the best one. This raises the need of extending  classic decision-theory  to take into account the available computational resources. 

Here we consider decision-makers with a prior distribution $p_0(x)$ over the search space and with limited computational resources. We measure computational resources in terms of an ``information distance'', namely the relative entropy $\dkl{p}{p_0} = \sum_x p(x) \log \frac{p(x)}{p_0(x)}$ between  the prior behaviour $p_0$ and a new behaviour $p$ after deliberation. This information-processing cost can be related to physical inefficiencies in thermodynamics~\cite{Ortega2013}. For every environment $y$ the decision problem can be phrased as a constrained maximization problem with $D_{\text{KL}} (p||p_0) \le B$, where the 
decision-maker can only afford to spend a maximum number of bits $B$ to change the prior behavior $p_0$ to the new behavior $p$, such that
\begin{equation}\label{eq:fe_objective}
\argmax_{p(x|y)} \feobj (y) = \argmax_{p(x|y)} \sum_x p(x|y)U(x, y) - \frac{1}{\beta} \dkl{p(x|y)}{p_0(x)}
\end{equation} 
where $1/\beta$ is the Lagrange multiplier and serves as a resource parameter that trades off utility and computational cost. For $\beta \rightarrow \infty$ we recover classic decision-theory, and for $\beta \rightarrow 0$ the decision-maker has no resources and acts according to the prior. 
We can extend the maximization problem in Eq.~\eqref{eq:fe_objective} to also maximize over the prior distribution in average over all environments
\begin{equation}\label{eq:rdis_objective}
\argmax_{p(x|y), p(x)} \rdisobj = \argmax_{p(x|y), p(x)} \sum_y p(y)  \left[ \sum_x p(x|y)U(x, y) - \frac{1}{\beta} \dkl{p(x|y)}{p(x)} \right]
\end{equation}
This objective function is  commonly known as the rate-distortion objective first stated by C. Shannon.
The solution to this problem that gives the optimal prior $p(x)$ is the following set of two analytic self-consistent equations
\begin{align}
p^*(x|y) &= \frac{1}{Z(y)}p(x)\exp(\beta U(x,y)) \label{eq:post_eq}\\
p(x) &= \sum_y p(y) p^*(x|y) \label{eq:prior_eq}
\end{align}
where $Z(y)= \sum_x p(x)e^{\beta U(x,y)}$ is the partition sum. In practice, the solution can be computed using the Blahut-Arimoto algorithm that basically iterates through both equations until convergence. However, every iteration includes the computation of the partition sum $Z(y)$ that can potentially be infeasible because it involves a function evaluation over all possible actions in the set $\setX$ which could be very large. Another inconvenience is that the Blahut-Arimoto algorithm can only be applied directly when the action space is discrete.

\section{Decision-maker with parametric prior} 
Here we propose a sampling-based approach that avoids the computation of partition sums and that can also be applied to continuous action spaces. In order to achieve this we require a parametrized family of distributions $p_\theta(x)$ to allow for a gradient ascent update rule. 
Replacing $p(x)$ in Eq.~\eqref{eq:post_eq}-\eqref{eq:prior_eq} by $p_\theta(x)$ and inserting into Eq.~\eqref{eq:rdis_objective}, we can rewrite the rate-distortion objective $\rdisobj$ as
\begin{equation}
\rdisobj (\theta) = \frac{1}{\beta }\sum_y p(y) \log Z_\theta(y) \nonumber
\end{equation}
where $Z_\theta(y) = \sum_x p_\theta (x) \exp (\beta U(x,y))$
and $p_\theta(x|y) = \frac{1}{Z_\theta(y)} p_\theta(x) \exp ( \beta U(x,y) )$.  The gradient  with respect to $\theta$ is 
\begin{align}
\gradtheta \rdisobj (\theta) =&  \frac{1}{\beta} \sum_y p(y) \frac{\sum_x p_\theta(x) \gradtheta \log p_\theta(x) \exp(\beta  U(x,y))}{ \sum_x p_\theta (x) \exp(\beta  U(x,y)} \nonumber \\
=& \frac{1}{\beta} \sum_y p(y) \sum_x p_\theta(x|y) \gradtheta \log p_\theta (x)
\label{eq:grad_final}
\end{align}
where in the first row we have used the equality $\gradtheta p_{\theta}(x) = p_\theta (x) \gradtheta \log p_\theta (x) $, and in the second row we used the definition of the parametric conditional $p_\theta(x|y)$. Due to the double expectation in Eq.~\eqref{eq:grad_final}, we can now approximate the gradient stochastically by  single samples $y' \sim p(y)$ and  $x' \sim p_\theta (x|y')$ and evaluating
$\gradtheta  \rdisobj (\theta) \approx \frac{1}{\beta} \gradtheta \log p_\theta (x')$.
The parameter updates are computed as usual with 
\begin{equation}
\label{eq:update}
\theta_{t+1}=\theta_t +  \frac{\alpha}{\beta} \gradtheta \log p_\theta (x')
\end{equation} where $\alpha$ is the learning rate. 
\paragraph{Sampling}
For the approximation $\gradtheta  \rdisobj (\theta) \approx \frac{1}{\beta} \gradtheta \log p_\theta (x')$ the decision-maker needs to generate a sample from $p_\theta (x|y')$ after being given the sample $y'$ from the environment.  
This can be achieved, for example, by a rejection sampling scheme. The rejection sampling algorithm takes samples from the proposal distribution $p_\theta(x)$ and accepts or rejects them according to the acceptance-rejection rule
\begin{equation}\label{eq:rejection_sampling}
u\le \min \left( 1, \frac{\exp (\beta U(x, y)}{\exp( \beta T(y) )}\right)
\end{equation}
where $u\sim U[0,1]$ is sampled from a uniform distribution and the aspiration level of the decision-maker for the environment $y$ is $T(y) \ge \max_{x} U(x, y)$.
If the actions $x$ correspond to accepted samples, it can be shown that the decision-maker effectively generates samples from the conditional $p_{\theta}(x|y)$. The average number of samples required for acceptance given a particular $y$ is
\begin{equation}
s(y) = \frac{\exp(\beta T(y))}{Z_\theta(y)} \ge \exp (\dkl{p_\theta (x|y)}{p_\theta(x)}) \nonumber
 \end{equation}
that is greater than the exponential of the relative entropy---connecting sampling complexity with the information-theoretic costs. The average number of required samples across all environments is naturally
\begin{equation}\label{eq:av_samplesallworlds}
S = \sum_y p(y) s(y).
\end{equation}
\section{Simulations}
\begin{figure}[t]
	\centering
	\includegraphics[scale=0.56]{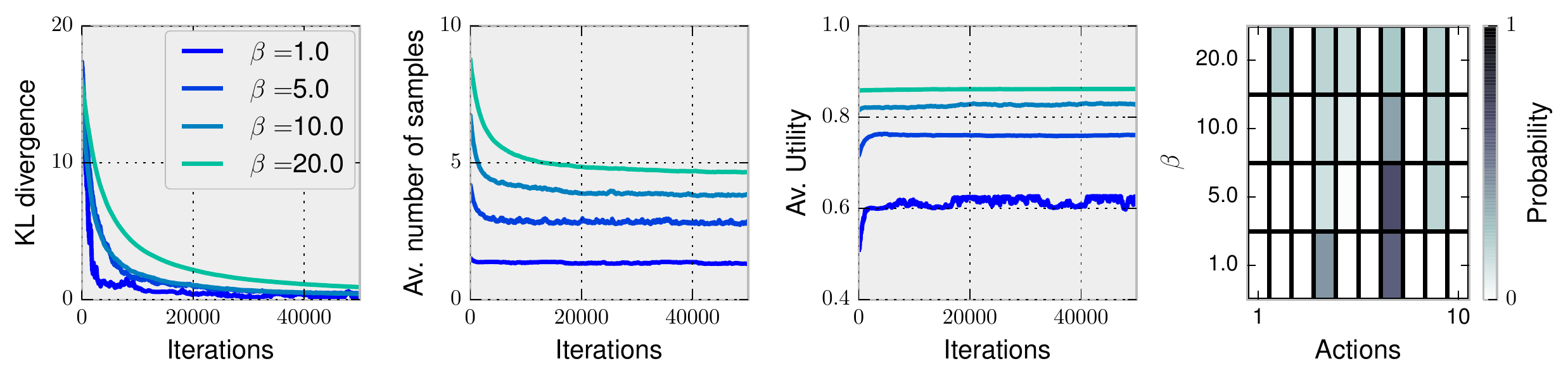}
	\caption{Experimental results with $\alpha = 0.05$ and different $\beta$ for $5$ different environments and $10$ possible actions. At every iteration we sample an environment $y' \sim p(y)$ and take an accepted action $x'\sim p_\theta(x|y)$ according to the rule of Eq.~\ref{eq:rejection_sampling}. From left to right we show the relative entropy between the optimal solution and $p_\theta (x)$, the average number of samples needed for acceptance, the average utility and finally the shape of the prior after convergence.  }
	\label{fig:results}
\end{figure}

While the proposed algorithm works for both continuous and discrete parametric distributions, here we demonstrate its performance in simulations for the discrete case, because this allows comparing the solutions to the optimal Blahut-Arimoto solutions. Additionally, these simulations illustrate the impact of $\beta$ on learning and plateau performance and on the average number of  samples required. 
\paragraph{The parametric prior} When using gradient methods in parameter space, it has to be made sure that the parameters do not violate given constraints---for example in the case of a discrete distribution over $n+1$ outcomes with parameters $\pbf = \left[p_0, p_1, \dots, p_n \right]$, the constraints $p_0=1-\sum_{i=1}^{n} p_i$ and $p_i \geq 0$ must be satisfied. Often the constraints can be naturally satisfied by reparameterizing the distribution. In the discrete case we may define actions for example as vectors of the form $\xbf_i = [ x_0, \dots x_i, \dots x_n ] $ with only one dimension having value $x_i = 1$ and the rest $x_j=0$ for all $j\neq i$ where $0\le i \le n$. Then we can reparameterize the discrete distribution as 
 \begin{equation}
p_{\thetabf}(\xbf ) = \exp\left( \sum_{i=1}^n \theta_i x_i - \psi (\thetabf) \right)
\end{equation}
with the new parameter vector $\thetabf = [ \theta_1 ,\dots \theta_n]$ and where $\psi (\thetabf) = \log \left( 1 + \sum_{i=1}^n \exp( \theta_i ) \right)$. The gradient $\gradtheta \log p_\theta (x)$ from Eq.~\eqref{eq:grad_final} can then be computed as
\begin{equation}
\nabla_{\thetabf} \log p_{\thetabf}(\xbf ) = 
\begin{bmatrix}
\frac{\partial}{\partial \theta_1} \log p_{\thetabf}(\xbf ) \\
\vdots \\
\frac{\partial}{\partial \theta_n} \log p_{\thetabf}(\xbf )  \\
\end{bmatrix}
=
\begin{bmatrix}
x_1 -\frac{\exp(\theta_1)}{1+ \sum \exp(\theta_i)}\\
\vdots\\
x_n -\frac{\exp(\theta_n)}{1+ \sum \exp(\theta_i)}\\
\end{bmatrix}.
\end{equation}
Note that with this reparameterization $\theta_i = \log \frac{p_i}{p_0}$ the parameters $\theta_i \in \mathbb{R}$ can vary across all reals, thereby naturally satisfying the constraints for $p_i$.
\paragraph{Simulations} In Fig.~\ref{fig:results} we show the evolution of a bounded rational decision-maker that updates the prior $p_\theta (x)$ according to Eq.~\eqref{eq:update} and uses the rejection sampling scheme from Eq.~\eqref{eq:rejection_sampling} to sample from the conditionals $p_\theta (x|y)$. For illustration we have chosen $|\setX| = 10$ and $|\setY| = 5$ and a utility function $U(x, y) \in [0;1]$ where the utility values for every $x$ and $y$ are sampled randomly between $0$ and $1$. 
The first panel shows how the sampling-based solutions converge to the optimal Blahut-Arimoto solutions, as the relative entropy between the parameterized prior $p_\theta (x)$ and the optimal prior (the solution to Eq.\eqref{eq:prior_eq}) is being reduced towards zero. It can be seen that the more rational decision-maker (higher $\beta$) takes longer to converge.
In the second panel, we show the average number of samples according to Eq. \eqref{eq:av_samplesallworlds}. It can be seen that for all decision-makers the number of required samples reduces over time as the prior gets closer to the optimal prior, indicating that the optimal solution is actually reducing the average information-theoretic distance between prior and conditionals. It can also be seen that more rational decision-makers require more samples. In the third panel, we show how the average utility $\sum_y p(y)\sum_x p_\theta(x|y) U(x,y) $ improves over time with a better prior. Finally, the fourth panel shows how the prior is shaped by the rationality parameter $\beta$. In a highly rational decision-maker the prior reflects an (more) evenly spread distribution between all the actions that are optimal in any of the environments. In a decision-maker with low rationality the prior will concentrate on the small set of actions that achieve on average a high utility across environments, because these decision-makers do not have resources to process environment-specific information.
\section{Conclusions}
Here we extend the information-theoretic model for bounded rational decision-making to allow for adaptation of the prior. This naturally leads to two time scales for information-processing. The first process is a slow process regarding the update of the prior distribution. We have shown that such an update can be estimated efficiently from samples and that it converges to the optimal solution obtained from rate distortion theory. The second process is the sampling process that a decision-maker with fixed prior would use to process information specific for a particular environment. A decision-maker with low rationality can be thought of as having no time for planning being forced to act according to the prior distribution that might not be a very good solution for the observed environment but that is the best possible prior given its computational resources.  Thus, this approach might not only provide a normative model for human decision-making processes occurring on different time scales, but also an efficient sampling-based solution for the rate distortion problem.
\paragraph{Acknowledgments} This study was supported by the DFG, Emmy Noether grant BR4164/1-1.
\small{

}

\end{document}